%% file: acl_latex.tex
\pdfoutput=1

\documentclass[11pt]{article}

\usepackage[]{ACL2023}

\usepackage{times}
\usepackage{latexsym}
\usepackage{graphicx}
\usepackage{float}
\usepackage{subfigure} 

\usepackage[T1]{fontenc}

\usepackage[utf8]{inputenc}

\usepackage{microtype}

\usepackage{array}
\usepackage{pifont}
\usepackage{lipsum}
\usepackage{tabularx}
\usepackage{adjustbox}
\usepackage{multirow}
\usepackage{enumitem}
\usepackage{xspace}
\usepackage{tcolorbox}
\usepackage{booktabs,amsfonts,dcolumn}
\usepackage{hyperref}
\usepackage{url}
\usepackage{amsmath,amsthm,amsfonts,amssymb,bm,stmaryrd,bbm}
\usepackage[noorphans,vskip=0.75ex,leftmargin=2ex]{quoting}
\usepackage{inconsolata}

\input{header}

\title{ \ours: Whitening-based Contrastive Learning of Sentence Embeddings}

\author{Wenjie Zhuo$^{1}$ \quad Yifan Sun$^{2}$ \quad Xiaohan Wang$^{1}$ \quad Linchao Zhu$^{1}$ \quad Yi Yang$^{1\dagger}$\\
  $^{1}$Zhejiang University, Hangzhou, China \\
  $^{2}$Baidu Inc., Beijing, China \\
  \texttt{\{12021057,yangyics\}@zju.edu.cn} \\
  \texttt{sunyf15@tsinghua.org.cn} \\
  \texttt{\{wxh1996111, zhulinchao7\}@gmail.com}}
  
\begin{document}
\maketitle

\renewcommand{\thefootnote}{\fnsymbol{footnote}}
\footnotetext{†Corresponding author.}
\renewcommand{\thefootnote}{\arabic{footnote}}

\input{sections/abstract.tex}

\input{sections/introduction.tex}

\input{sections/relate_work.tex}

\input{sections/methods.tex}

\input{sections/experiments.tex}

\input{sections/analysis.tex}
\input{sections/conclusion.tex}
\input{sections/limitation.tex}




\bibliography{anthology,custom}
\bibliographystyle{acl_natbib}

\end{document}

%% file: header.tex
\newcommand\ti[1]{\textit{#1}}

\newcommand\tf[1]{\textbf{#1}}
\newcommand\ttt[1]{\texttt{#1}}
\newcommand\mf[1]{\mathbf{#1}}

\newcommand\mr[1]{\mathrm{#1}}

\newcommand{\cls}{\ttt{[CLS]}}

\newcommand{\ours}{WhitenedCSE\xspace}

\newcommand{\la}{$_\texttt{large}$}
\newcommand{\ba}{$_\texttt{base}$}

\renewcommand{\paragraph}[1]{\vspace{0.2cm}\noindent\textbf{#1}}

%% file: sections/abstract.tex
\begin{abstract}

This paper presents a whitening-based contrastive learning method for sentence embedding learning (WhitenedCSE), which combines contrastive learning with a novel shuffled group whitening. Generally, contrastive learning pulls distortions of a single sample (\emph{i.e.}, positive samples) close and push negative samples far away, correspondingly facilitating the alignment and uniformity in the feature space. A popular alternative to the ``pushing'' operation is whitening the feature space, which scatters all the samples for uniformity. Since the whitening and the contrastive learning have large redundancy \emph{w.r.t.} the uniformity, they are usually used separately and do not easily work together. For the first time, this paper integrates whitening into the contrastive learning scheme and facilitates two benefits. \textbf{1}) Better uniformity. We find that these two approaches are not totally redundant but actually have some complementarity due to different uniformity mechanism. 
\textbf{2}) Better alignment. We randomly divide the feature into multiple groups along the channel axis and perform whitening independently within each group. By shuffling the group division, we derive multiple distortions of a single sample and thus increase the positive sample diversity. Consequently, using multiple positive samples with enhanced diversity further improves contrastive learning due to better alignment. 
Extensive experiments on seven semantic textual similarity tasks show our method achieves consistent improvement over the contrastive learning baseline and sets new states of the art, \emph{e.g.}, 78.78\% (+2.53\% based on BERT\ba) Spearman correlation on STS tasks.\footnote{Our code will be available at \url{https://github.com/SupstarZh/WhitenedCSE}.}

\end{abstract}

%% file: sections/introduction.tex
\section{Introduction}
\input{figures/vis_distribution.tex}
This paper considers self-supervised sentence representation (embedding) learning.        
It is a fundamental task in language processing (NLP) and can benefit a wide range of downstream tasks~\cite{qiao2016less,le2014distributed,lan2019albert,logeswaran2018efficient}.
Two characteristics matter for sentence embeddings, \emph{i.e.}, uniformity (of the overall feature distribution) and alignment (of the positive samples), according to a common sense in deep representation learning \cite{wang2020understanding}. Alignment expects minimal distance between positive pairs, while uniformity expects the features are uniformly distributed in the representation space in overall. From this viewpoint, the popular masked language modeling (MLM)~\cite{devlin2018bert,liu2019roberta,brown2020language,reimers2019sentence} is not an optimal choice for sentence embedding: MLM methods do not explicitly enforce the objective of uniformity and alignment and thus do not quite fit the objective of sentence representation learning. 

To improve the uniformity as well as the alignment, there are two popular approaches, \emph{i.e.}, contrastive learning and post-processing. 1) The contrastive learning methods~\cite{yan2021consert, gao2021simcse, kim2021self, wang2021cline} pulls similar sentences close to each other and pushes dissimilar sentences far-away in the latent feature space. 
Pulling similar sentences close directly enforces alignment, while pushing dissimilar sentences apart implicitly enforces uniformity \cite{wang2020understanding}. 
2) In contrast, the post-processing methods mainly focus on improving the uniformity. They use normalizing flows~\cite{li2020sentence} or whitening operation~\cite{su2021whitening}) to project the already-learned representations into an isotropic space. In other words, these methods scatter all the samples into the feature space and thus improve the uniformity. 

In this paper, we propose a whitening-based contrastive learning method for sentence representation learning (WhitenedCSE). For the first time, we integrate whitening into the contrastive learning scheme and demonstrate substantial improvement. Specifically, WhitenedCSE combines contrastive learning with a novel Shuffled Group Whitening (SGW). Given a backbone feature, SGW randomly divides the feature into multiple groups along the channel axis and perform whitening independently within each group. The whitened features are then fed into the contrastive loss for optimization.

Although the canonical whitening (or group whitening) is only beneficial for uniformity, SGW in WhitenedCSE improves not only the uniformity but also the alignment. We explain these two benefits in details as below:

$\bullet$ Better uniformity. 
We notice that the pushing effect in contrastive learning and the scattering effect in the whitening have large redundancy to each other, because they both facilitate the uniformity. This redundancy is arguably the reason why no prior literature tries to combine them. Under this background, our finding \emph{i.e.}, these two approaches are not totally redundant but actually have some complementarity is non-trivial. We think such complemenetarity is because these two approaches have different uniformity mechanism and will discuss the differences in Section \ref{sec:connect_to_cl}. In Fig.~\ref{fig:vis_dis}, we observe that while the contrastive learning (Fig.~\ref{fig:vis_dis} (b)) already improves the uniformity over the original bert features (Fig.~\ref{fig:vis_dis} (a)), applying whitening (Fig.~\ref{fig:vis_dis} (c)) brings another round of uniformity improvement. 

$\bullet$ Better alignment.
In the proposed WhitenedCSE, SGW is featured for its shuffled grouping operation, \emph{i.e.}, randomly dividing a backbone feature into multiple groups before whitening. Therefore, given a same backbone feature, we may repeat SGW multiple times to get different grouping results, and then different whitened features. These ``duplicated'' features are different from each other and thus increase the diversity of positive samples, as shown in Fig.~\ref{fig:vis_dis} (d). Using these diverse positive samples for contrastive learning, WhitenedCSE improves the alignment. 

Another important advantage of SGW is: since it is applied onto the backbone features, it incurs very slight computational overhead for generating additional positive samples. This high efficiency allows WhitenedCSE to increase the number of positive samples (more than common setting of 2) in a mini-batch with little cost. Ablation study shows that the enlarged positive-sample number brings a further benefit.

Our contributions are summarized as follows:

(1) We propose \ours for the self-supervised sentence representation learning task. \ours combines the contrastive learning with a novel Shuffled Group Whitening (SGW). 

(2) We show that through SGW, \ours improves not only the uniformity but also the alignment. Moreover, SGW enables efficient multi-positive training, which is also beneficial.

(3) We evaluate our method on seven semantic textual similarity tasks and seven transfer tasks. Experimental results show that \ours brings consistent improvement over the contrastive learning baseline and sets new states of the art.

%% file: figures/vis_distribution.tex
\begin{figure}[h]
    \centering 
    \subfigure[bert]{
        \label{subfig:1}
        \includegraphics[width=0.45\columnwidth]{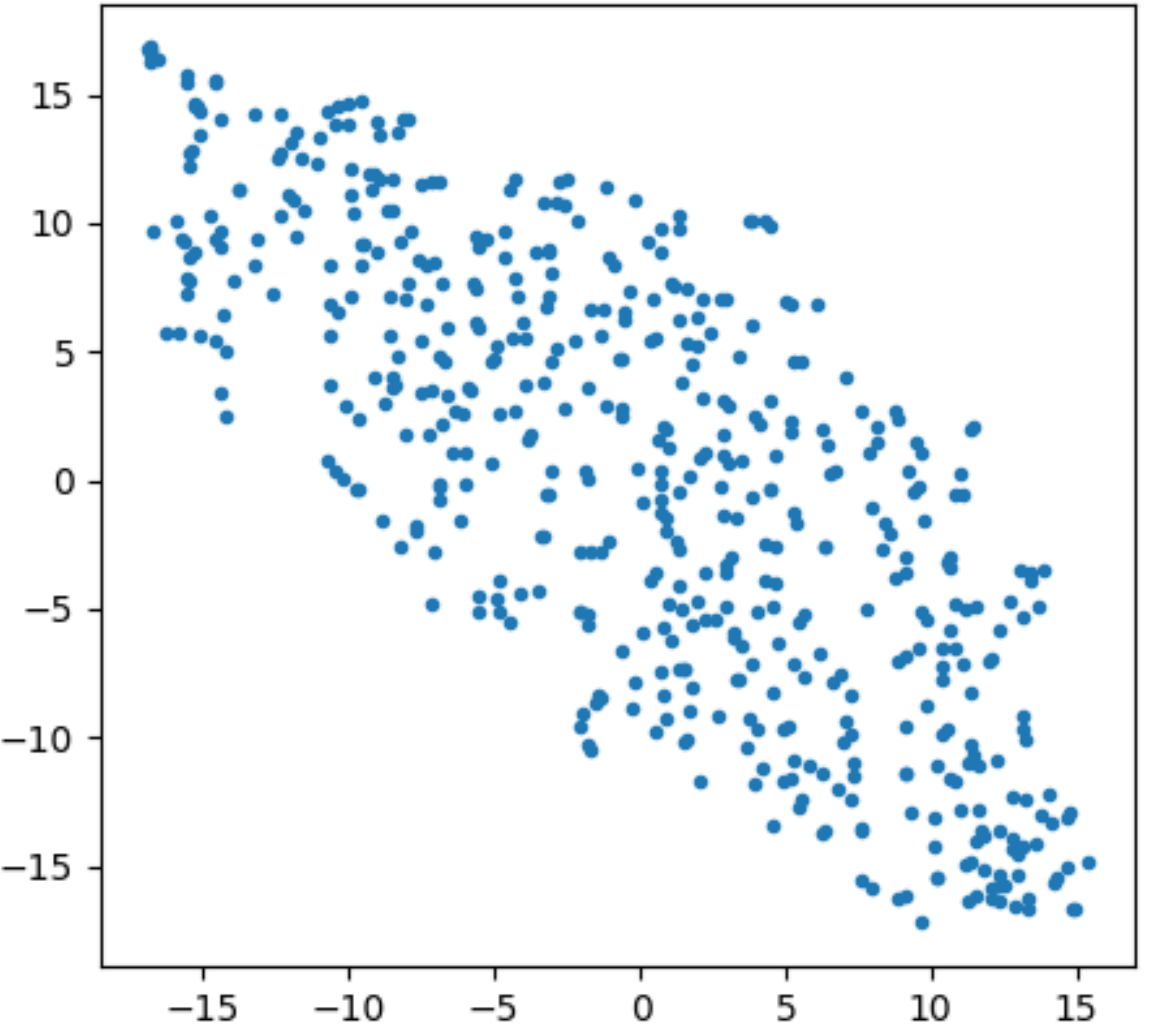}
    }\subfigure[bert+contrastive]{
        \label{subfig:2}
        \includegraphics[width=0.45\columnwidth]{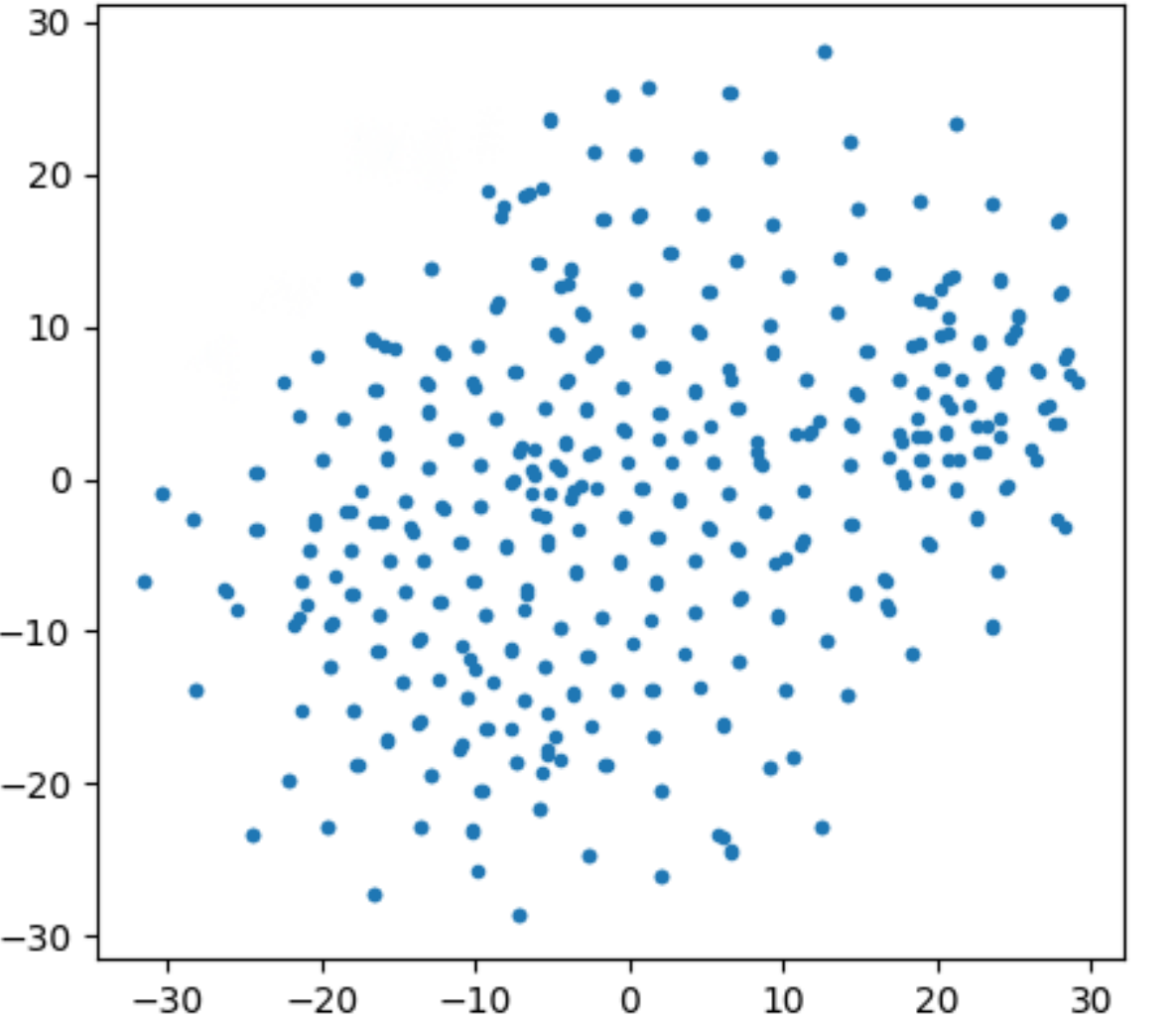}
    }
    \subfigure[bert+SGW+contrastive]{
        \label{subfig:3}
        \includegraphics[width=0.45\columnwidth]{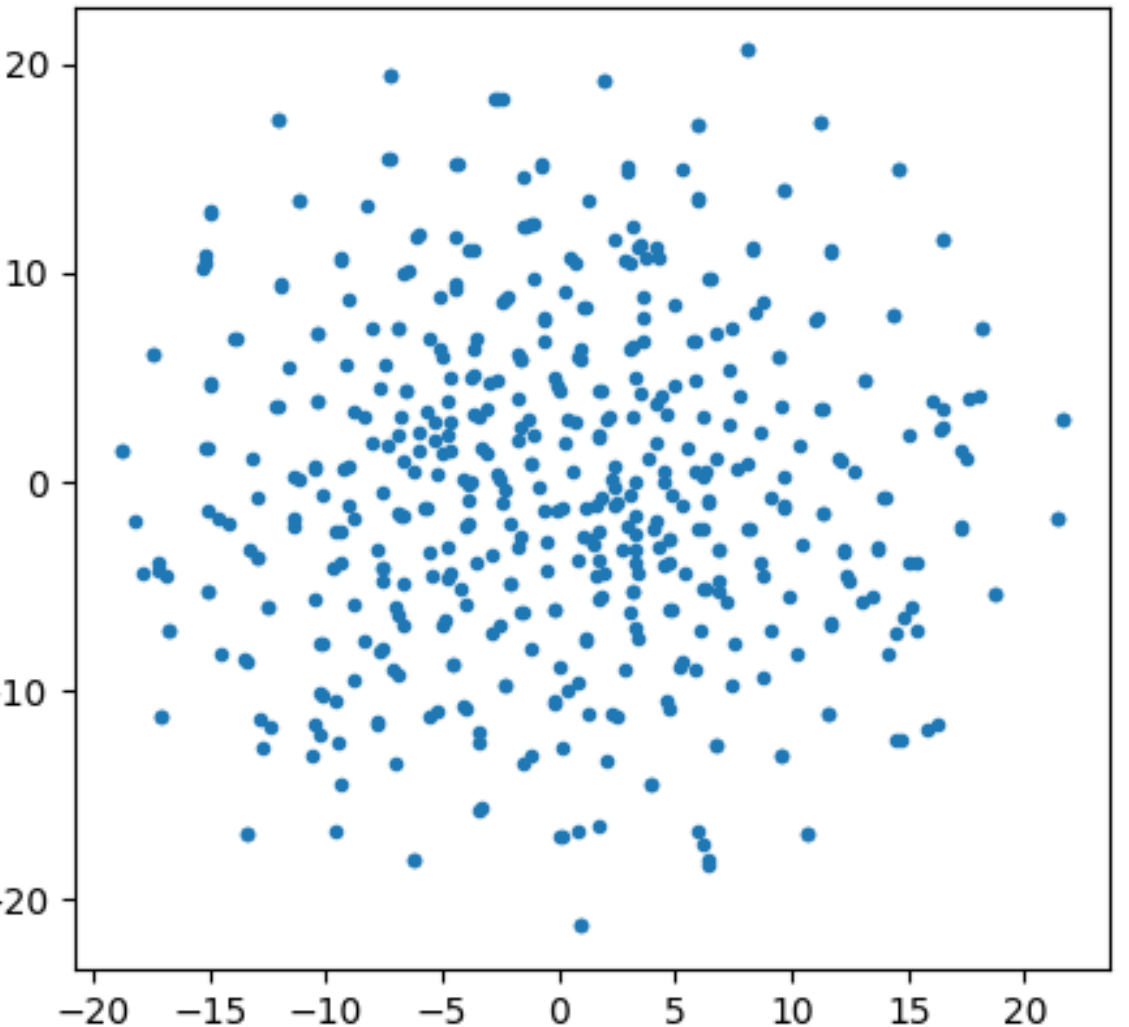}
    }\subfigure[positive distribution]{
        \label{subfig:4}
        \includegraphics[width=0.45\columnwidth]{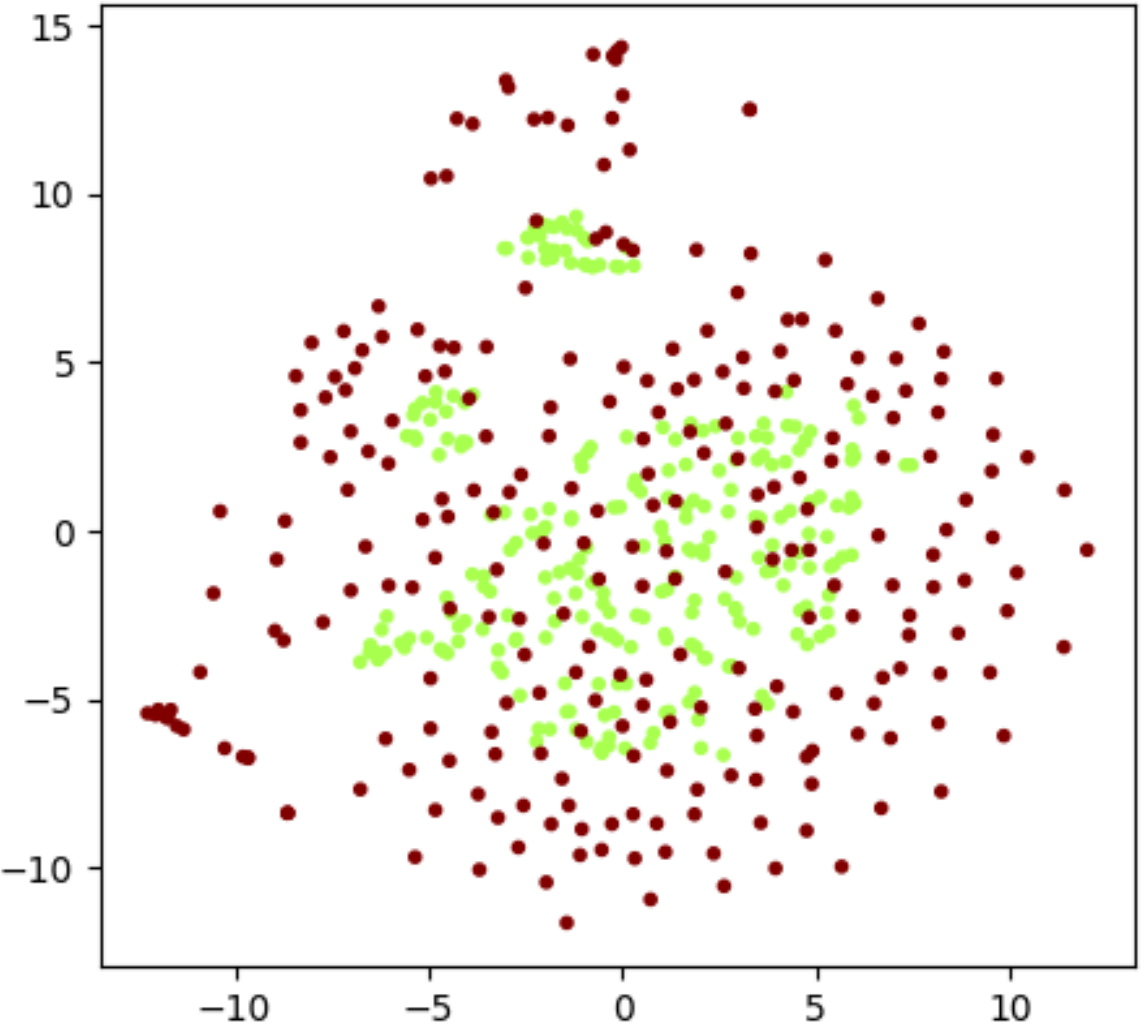}
    }
    \caption{
    The uniformity gradually improves in the deep embedding of (a) bert, (b) bert + contrastive, and (c) bert + SGW + contrastive, \emph{i.e.}, the proposed WhitenedCSE. Meanwhile, in (d), the positive samples after SGW (\textcolor{red}{red}) obtain higher diversity than the original bert features (\textcolor{green}{green}). Using these diverse positive samples for contrastive learning, the proposed \ours achieves better alignment.}
    
    \label{fig:vis_dis}
\end{figure}

%% file: sections/relate_work.tex
\section{Related Work}

\subsection{Sentence Representation Learning}
As a fundamental task in natural language processing, sentence representation learning has been extensively studied. Early works mainly based on bag-of-words~\cite{wu2010semantics,tsai2012bag} or context prediction tasks~\cite{kiros2015skip,hill2016learning}, \emph{etc}. Recently, with the advent of pretrained language model~\cite{devlin2018bert, liu2019roberta, 2020Language}, many works tend to directly use PLMs, such as BERT~\cite{devlin2018bert}, to generate sentence representations. However, some studies~\cite{ethayarajh2019contextual,yan2021consert} found that directly use the [CLS] representation or the average pooling of token embeddings at the last layer will suffer from \emph{anisotropy} problem, \emph{i.e.}, the learned embeddings are collasped into a small area. To alleviate this problem, BERT-flow~\cite{li2020sentence} adopts a standardized flow transformation while BERT-Whitening~\cite{su2021whitening} adopts a whitening transformation, both of them transform the representation space to a smooth and isotropic space. Most recently, contrastive learning~\cite{chen2020simple, gao2021simcse} has become a powerful tool to obtain the sentence representations.

\subsection{Contrastive Learning}
Contrastive learning~\cite{chen2020simple,2020Momentum} has achieved great success in sentence representation learning tasks~\cite{gao2021simcse, yan2021consert, kim2021self, wang2021cline}. It pulls semantically similar samples together, and pushes the dissimilar samples away, which can be formulated as:
\begin{equation}
    \label{eq:contrastive_loss}
    \begin{aligned}
        \mathcal{L}_{cl} = -log \frac{e^{sim(h_i, h_i^*)/\tau}}{\sum_{j=1}^ne^{sim(h_i,h_j^*)/\tau}}
    \end{aligned}
\end{equation}

where $\tau$ is a temperature hyperparameter, $\mf{h_i^*}$, $\mf{h_j^*}$ are the positive sample and negative samples respectively. 
Recently, \ti{alignment} and \ti{uniformity}~\cite{wang2020understanding} are proposed to measure the quality of representations. Alignment measures whether the distance between positive samples is close, while uniformity measures the dispersion of embedding in vector space. A typical method called SimCSE~\cite{gao2021simcse} uses dropout as a feature-wise data augmentation to construct the positive sample, and randomly sample negatives from the batch, which can achieve a great balance between alignment and uniformity. Some new works further improved the quality of sentence representations based on SimCSE, such as ESimCSE~\cite{wu2021esimcse}, MixCSE~\cite{zhang2022unsupervised} and VaSCL~\cite{zhang2021virtual}, each of them proposed a new data augmentation strategy to construct the positive pair. Besides, DCLR~\cite{zhou2022debiased} focus on optimizing the strategy of sampling negatives, and ArcCSE~\cite{zhang2022contrastive} optimized the objective function, \emph{etc}.
In this paper, we find that contrastive learning can be further combined with whitening to obtain better sentence representations.



\subsection{Whitening Transformation}
In computer vision, recent works~\cite{ermolov2021whitening, zhang2021zero, hua2021feature} use whitening transformation as an alternative method to the "pushing negatives away" operation in contrastive learning to disperse the data uniformly throughout the spherical space (\emph{i.e.,} the feature space), and then pull the positive samples together , which have achieved great success in unsupervised representation learning. 

Whitening (\emph{aka.,} sphering) is a common transformation that transforms a set of variables into a new set of isotropic variables, and makes the covariance matrix of whitened variables equal to the identity matrix. In natural language processing, ~\citet{su2021whitening} use whitening as a post-processing method to alleviate the anisotropic problem in pretrained language models.
In this paper, we use whitening as an explicit operation to further improve the uniformity of the representation space, and further explore the potential of whitening in improving alignment, so as to obtain a better sentence representation model.

%% file: sections/methods.tex
\section{Methods}
In this section, we first describe the overall architecture of \ours and then present the details of all the modules, including the shuffled group whitening module and the new contrastive learning module. 


\input{figures/main_figure.tex}
\subsection{General Framework}
As shown in Fig.~\ref{fig:structure}, \ours has three major components: 

$\bullet$
An BERT-like encoder
, which we use to extract features from native sentences, and take the [CLS] token as our native sentence representations. 

$\bullet$
Shuffled-group-whitening module, we use it as a complementary module to contrastive learning to further improve the uniformity and alignment of the representation space. 

$\bullet$
Multi-positives contrastive module, in this module, we pull distortions of the representations close and push the negative samples away in the latent feature space.

Specifically, given a batch of sentences $\mf{X}$, \ours use the feature encoder $f_{\theta}(x_i, \gamma)$ to map them to a higher dimensional space, where $\gamma$ is a random mask for dropout~\cite{gao2021simcse}, then we take the [CLS] output as the native sentence representations. After this, we feed the native sentence representations to the shuffled-group-whitening (SGW) module, in this module we randomly dividing each sentence representation into multiple groups along the axis, then we operate group whitening on each group. We repeat SGW multiple times to get different grouping results, and then different whitened representations. These "duplicated" features are different from each other. Finally, we use multi-positives contrastive loss function to pull one representation and all its corresponding augmentations together, and push it away from others. We will discuss feasible loss function in Section \ref{sec:loss_function}, and present our final form of loss function.
\subsection{From Whitening to SGW}
\label{sec:whitening}
\subsubsection{Preliminaries for Whitening}
Given a batch of normalized sentence representations $\mf{Z} \in \mathbb{R}^{N \times d}$, the whitening transformation can be formulated as:
\begin{equation}
    \label{eq:whitening}
    \begin{aligned}
        \mathbf{H} = \mathbf{Z}^{\mathrm{T}}\mathbf{W}
    \end{aligned}
\end{equation}

where $\mf{H} \in \mathbb{R}^{d \times N}$ is the whitened embeddings and $\mf{W} \in \mathbb{R}^{d \times d}$ is the whitening matrix. We denote the covariance matrix of $\mf{ZZ^{\mathrm{T}}}$ as $\Sigma$. the goal of whitening is to make the covariance matrix of $\mf{HH^{\mathrm{T}}}$ equal to the identity matrix $\mf{I}$, \emph{i.e.}, $\mf{W}\Sigma\mf{W^{\mathrm{T} }} = \mf{I}$. There are many different whitening methods, such as PCA~\cite{jegou2012negative}, ZCA~\cite{bell1997independent}, etc. Group whitening use ZCA as its whitening method to prevent the \ti{stochastic axis swapping}~\cite{huang2018decorrelated}.\footnote{\ti{stochastic axis swapping} can drastically change the data representation from one batch to another such that training never converges~\cite{huang2018decorrelated}.}.


\paragraph{ZCA Whitening.} 
The whitening matrix of ZCA whitening transformation can be formulated as:
\begin{equation}
    \label{eq:ZCA}
    \begin{aligned}
        \mathbf{W}^{ZCA} = \mathbf{U \Lambda^{-1/2} U^{\mathrm{T}}}
    \end{aligned}
\end{equation}
where $\mf{U} \in \mathbb{R}^{d \times d}$ is the stack of eigenvector of $\ti{cov}\mf{(Z, Z^{\mathrm{T}})}$, and $\mf{\Lambda}$ is the correspond eigenvalue matrix. $\mf{U}$ and $\mf{\Lambda}$ are obtained by matrix decomposition. Therefore, Eq.~\ref{eq:whitening} becomes:
\begin{equation}
    \label{eq:ZCA_whitening}
    \begin{aligned}
        \mathbf{H} =  \mathbf{Z}^{\mathrm{T}} \mathbf{U \Lambda^{-1/2} U^{\mathrm{T}}}
    \end{aligned}
\end{equation}

\paragraph{Group Whitening.} 
Since whitening module needs a large batch size to obtain a suitable estimate for the full covariance matrix, while in NLP, large batch size can be detrimental to unsupervised contrastive learning. To address this problem, we use group whitening~\cite{huang2018decorrelated}, which controls the extent of whitening by decorrelating smaller groups. Specifically, give a sentence representation of dimension $d$,  group whitening first divide it into $k$ groups $\mf{(Z_0, Z_1, ..., Z_{k-1})}$, i.e., $\mf{Z_k} \in \mathbb{R}^{N \times \frac{d}{k}}$ and then apply whitening on each group. That is:
\begin{equation}
    \label{eq:Group_whitening}
    \begin{aligned}
        \mathbf{H} = \ti{concat}(\mathbf{Z}_{i}\mathbf{W}^{ZCA}_{i}),~ i \in [0, k)    
    \end{aligned}
\end{equation}

\subsubsection{Shuffled Group Whitening}
In order to further improve the quality of the sentence representation model, we proposed shuffled-group-whitening (SGW). We randomly divide the feature into multiple groups along the channel axis, and then perform ZCA whitening independently within each group. After whitening, we do a re-shuffled operation to recover features to their original arrangement. The process can be formulated as:
\begin{equation}
    \label{eq:shuffled}
    \begin{aligned}
        \mathbf{H} = \ti{shuffled}^{-1}(\mathbf{GW}(\ti{shuffled}(\mathbf{Z}^{\mathrm{T}})))
    \end{aligned}
\end{equation}

This can bring two benefits. One is that it can avoid the limitation that only adjacent features can be put into the same group, so as to better decorrelation and then achieve better uniformity in the representation space. Another is that it brings a disturbance to samples, we can use it as a data augmentation method. Specifically, we repeat SGW multiple times and we can get different grouping results and then different whitened features. These "duplicated" features are different from each other, and thus increase the diversity of postive samples.

\subsubsection{Connection to contrastive learning}
\label{sec:connect_to_cl}
We find that whitening and contrastive learning are not totally redundant but actually have some complementarity is non-trivia. Specifically, whitening decorrelates features through matrix decomposition, and makes the variance of all features equal to 1, that is, to project features into a spherical space. The "pushing" operation in contrastive learning is to approach a uniform spherical spatial distribution step by step through learning/iteration. Therefore, conceptually, whitening and contrastive learning are redundant in optimizing the uniformity of the representation space.
However, contrastive learning achieves uniformity by widening the distance between positive samples and all negative samples, but there is no explicit separation between negative samples. Whitening is the uniform dispersion of the entire samples, so there is complementarity between them. That is, whitening can supplement the lack of contrastive learning for the "pushing" operation between negative samples.

\subsection{Multi-Positive Contrastive Loss}
\label{sec:loss_function}
Since we get multi-positive samples from SGW module, however, the original contrastive loss in Eq.~\ref{eq:contrastive_loss} is unable to handle multiple positives. We provide two possible options of contrastive loss which can adapt multi-positives. Given $\mf{m}$ positive samples, the objective function can be formulated as:
\begin{equation}
    \label{eq:multi_pos1}
    \begin{aligned}
       \mathcal{L}_1 =  -\lambda_m \sum_{p=1}^{m} \log \frac{e^{\mr{sim}(\mf{h}_i, \mf{h}^+_{i,p})/\tau}}{\sum_{j=1}^N e^{\mr{sim}(\mf{h}_i, \mf{h}_{j}^+)/\tau}}
    \end{aligned}
\end{equation}


\begin{equation}
    \label{eq:multi_pos2}
    \begin{aligned}
       \mathcal{L}_2 =  - \log  \sum_{p=1}^{m} \frac{\lambda_m e^{-\mr{sim}(\mf{h}_i, \mf{h}^+_{i,p})/\tau}}{\sum_{j=1}^N e^{\mr{sim}(\mf{h}_i, \mf{h}_{j}^+)/\tau}}
    \end{aligned}
\end{equation}

where $\lambda_m$ is a hyperparameter, it controls the impact of each positive. Eq.~\ref{eq:multi_pos1} puts the summation over positives outside of the log while Eq.~\ref{eq:multi_pos2} puts the sum of positives inside the log. It should be noted that in Eq.~\ref{eq:multi_pos2}, there is a negative sign before the sum of positives, without it, the Eq.~\ref{eq:multi_pos2} will conduct hard mining, which means the maximum of $\sum_{p=1}^{m}e^{\mr{sim}(\mf{h}_i, \mf{h}^+_{i,p})/\tau}$ is mainly determined by $max(e^{-\mr{sim}(\mf{h}_i, \mf{h}^+_{i,p})/\tau})$. If we add the negative sign, the loss function will be committed to punish the items with less similarity, which is good for bringing all positive samples closer to the anchor samples. In our framework, we adopt Eq.~\ref{eq:multi_pos1} as our final loss function because it can achieve better performance.

%% file: figures/main_figure.tex
\begin{figure*}[!ht]
    \centering
    \includegraphics[width=0.98\textwidth]{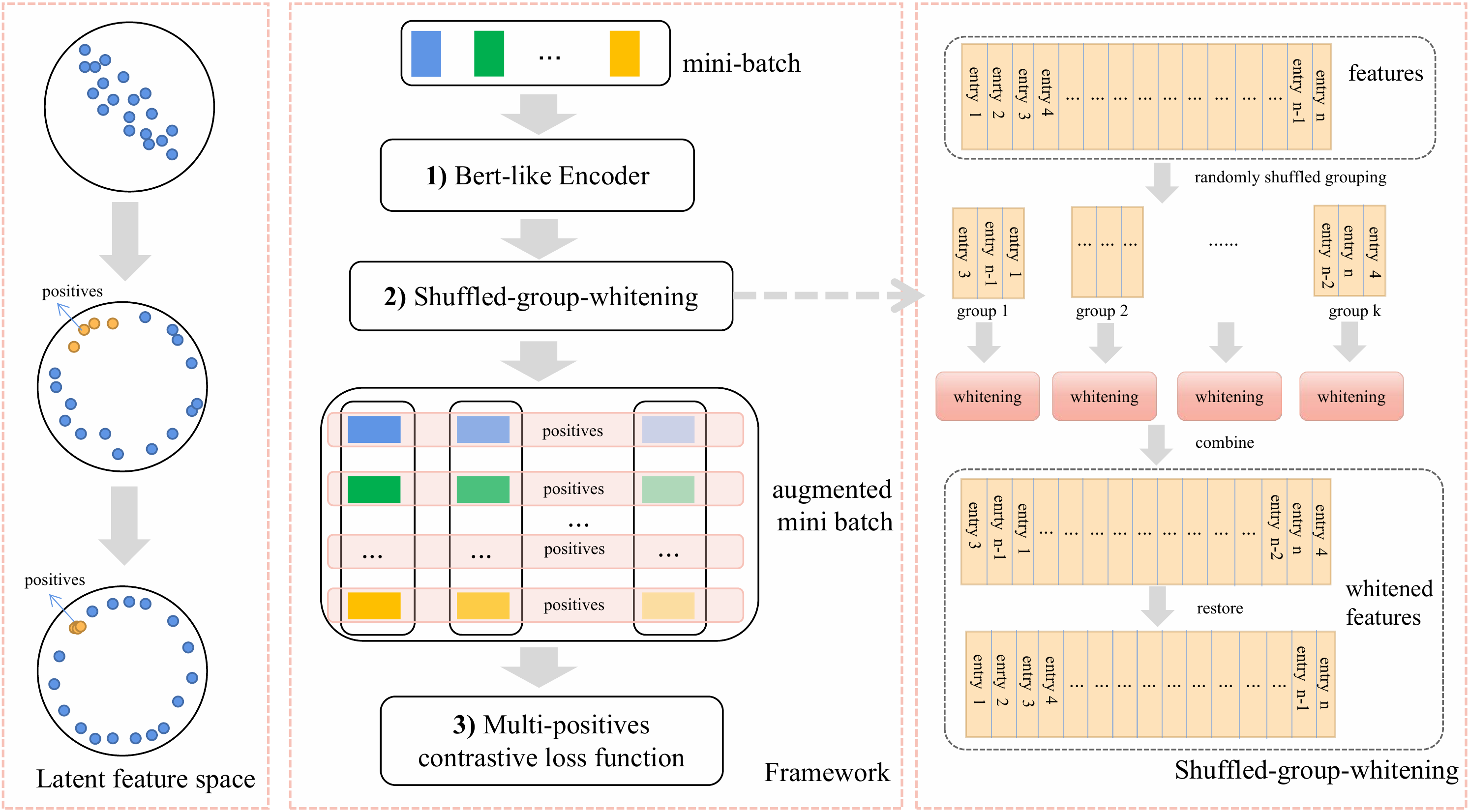}
    \caption{
    Method overview. In the middle column, \ours consists of three components, \emph{i.e.}, 1) An BERT-like encoder for generating the backbone features from input samples, 2) A Shuffled Group Whitenning (SGW) module for scattering the backbone features and augmenting the positive feature diversity, and 3) A multi-positive contrastive loss for optimizing the features. In the left column, when the mini-batch flows through these three components sequentially, the feature space undergoes ``anisotropy'' $\rightarrow$ ``good uniformity + augmented positives'' $\rightarrow$ ``pulling close the positives''. The right column illustrates the SGW module in details. Specifically, SGW randomly shuffles the backbone feature along the axis and then divides the feature into multiple groups. Afterwards, SGW whitens each group independently and re-shuffles the whitened feature. Given a single backbone feature, we repeat the SGW process several times so as to generate multiple positive features. }
    

    \label{fig:structure}
\end{figure*}

%% file: sections/experiments.tex
\section{Experiments}
\label{sec:experiment}
\input{tables/main_sts}
\input{tables/main_transfer}

\subsection{Experiment Setup}
In this section, We evaluate our method on seven Semantic Textual Similarity(STS) tasks and seven transfer tasks. We use the SentEval~\cite{conneau2018senteval} toolkit for all of tasks.

\paragraph{Datasets.} 
Semantic Textual Similarity(STS) tasks consist of seven tasks: STS 2012--2016~\cite{agirre-etal-2012-semeval,agirre-etal-2013-sem,agirre-etal-2014-semeval,agirre-etal-2015-semeval,agirre-etal-2016-semeval}, STS Benchmark~\cite{cer-etal-2017-semeval} and SICK-Relatedness~\cite{marelli-etal-2014-sick}. Each sample in those datasets has two sentences and a manually annotated similarity score from 0 to 5 to measure their similarity. The transfer tasks include MR~\cite{pang2005seeing_mr}, CR~\cite{hu2004mining_cr}, SUBJ~\cite{pang2004sentimental_subj}, MPQA~\cite{wiebe2005annotating_mpqa}, SST-2~\cite{socher2013recursive_sst-2}, TREC~\cite{voorhees2000building_trec}, MRPC~\cite{dolan-brockett-2005-automatically-mrpc}. In this tasks, we use a logistic regression classifier trained on top of the frozen sentence embeddings.

\paragraph{Baseline and competing methods.}
We compare \ours against several classic methods on Semantic Textual Similarity datasets, \emph{i.e.}, GloVe embeddings~\cite{pennington2014glove}, average BERT embeddings from the last layer~\cite{devlin2018bert}, BERT-flow~\cite{li2020sentence}, BERT-whitening~\cite{su2021whitening}, IS-BERT~\cite{zhang2020unsupervised}, CT~\cite{carlsson2021amaru}, ConSERT~\cite{yan2021consert}, SimCSE~\cite{gao2021simcse}, as well as some most recent state-of-the-art methods, \emph{i.e.}, MixCSE~\cite{zhang2022unsupervised}, ArcCSE~\cite{zhang2021pairwise}, DCLR~\cite{zhou2022debiased}.

Among these methods, SimCSE may be viewed as our direct baseline, because \ours may be viewed as being transformed from SimCSE by adding the SGW and replacing the dual-positive contrastive loss with multi-positive contrastive loss. Therefore, when conduct ablation study, we use SimCSE as our baseline. 

\paragraph{Implementation details.} 
We use the output of the MLP layer on top of the [CLS] as the our sentence representation. The MLP layer is consist of three components, which are a shuffled group whitening module, a 768 $\times$ 768 linear layer and a activation layer. Following SimCSE~\cite{gao2021simcse}, we use $1 \times 10^6$ randomly sampled sentences from English Wikipedia as our training corpus. We start from pre-trained checkpoints of BERT~\cite{devlin2018bert} and RoBERTa~\cite{liu2019roberta}. At training time, we set the learning rate as 3e-5, the batch size as 64. We train our model for 1 epoch with temperature $\tau=0.05$. For BERT-base and BERT-large, we set the number of group size as 384, for RoBERTa-base and RoBERTa-large, we set the number of group size as 256. We set the number of positives as 3 for all of models. We evaluate the model every 125 training steps on the development set of STS-B, and keep the best checkpoint for evaluation on test sets. We conduct our experiments on two 3090 GPUs.


\subsection{STS tasks}
We conduct experiments on 7 semantic textual similarity(STS) tasks, and use SentEval toolkit~\cite{conneau2018senteval} for evaluation. We use the Spearman's correlation coefficient as our evaluation metrics. The Spearman's correlation uses a monotonic equation to evaluate the correlation of two statistical variables, it varies between -1 and 1 with 0 implying no correlation, and the closer the value is to 1, the closer the two statistical variables are to positive correlation.
    

Tab.~\ref{tab:main_sts} shows the evaluation results on 7 STS tasks, from which we can see that \ours achieves competitive performance. Compared with SimCSE~\cite{gao2021simcse}, \ours achieves 2.53 and 1.56 points of improvement based on BERT\ba and BERT\la. It also raise the performance from 76.57\% to 78.22 \% base on RoBERTa\ba. Compared with recent works, \ours also achieves the best performance in most of the STS tasks.

\subsection{Transfer tasks}
We also conduct experiments on 7 transfer tasks, and use SentEval toolkit~\cite{conneau2018senteval} for evaluation. For each task, we train a logistic regression classifier on top of the frozen sentence embeddings and test the accuracy on the downstream task. In our experiment settings, we do not include models with auxiliary tasks, i.e., masked language modeling, for a fair comparison.

Tab.~\ref{tab:main_transfer} shows the evaluation results. Comparied with the SimCSE~\cite{gao2021simcse} baseline, \ours achieves 0.59 and 0.33 accuracy improvement on average results based on BERT\ba and BERT\la. Compared with recent works, \ours also achieves the best performance in most of the transfer tasks, which further demonstrates the effectiveness of our method.

\input{figures/alignment_and_uniformity.tex}
\subsection{Alignment and Uniformity}
In order to further quantify the improvement in uniformity and alignment of \ours, we follow SimCSE~\cite{gao2021simcse}, and use alignment loss and uniformity loss~\cite{wang2020understanding} to measure the quality of representations. Alignment is used to measure the expected distance between the embeddings of the positive pairs, and can be formulated as:
\begin{equation}
    \label{eq:alignment}
    \resizebox{.73\hsize}{!}{%
    $
    \ell_{\mr{align}}=\underset{(x, x^+)\sim p_{\mr{pos}}}{\mathbb{E}} \Vert f(x) - f(x^+) \Vert^2
    $
    }
\end{equation}
while uniformity measures how well the embeddings are uniformly distributed in the representation space:
\begin{equation}
    \resizebox{.85\hsize}{!}{%
    $
    \label{eq:uniformity}
    \ell_{\mr{uniform}}=\log \underset{~~~x, y\stackrel{i.i.d.}{\sim} p_{\mr{data}}}{\mathbb{E}}   e^{-2\Vert f(x)-f(y) \Vert^2}
    $
    }
\end{equation}

We calculate the alignment loss and uniformity loss every 125 training steps on the STS-B development set. From Fig.~\ref{fig:alignment_and_uniformity}, we can see that  compared with SimCSE, \ours performs better both on the alignment measure and the uniformity measure. We also find that the uniformity of our models is well optimized at the beginning and remains stable throughout the training process. This further confirms that our method can improve the quality of sentence representation more effectively.

%% file: tables/main_sts.tex
\begin{table*}[t]
    \begin{center}
    \centering
    \small
    \begin{tabular}{lcccccccc}
    \toprule
       \tf{Model} & \tf{STS12} & \tf{STS13} & \tf{STS14} & \tf{STS15} & \tf{STS16} & \tf{STS-B} & \tf{SICK-R} & \tf{Avg.} \\
    \midrule
    \midrule
        GloVe embeddings (avg.) & 55.14 & 70.66 & 59.73 & 68.25 & 63.66 & 58.02 & 53.76 & 61.32 \\
        BERT\ba~(first-last avg.) & 39.70&	59.38&	49.67&	66.03&	66.19&	53.87&	62.06&	56.70\\ 
        BERT\ba-flow & 58.40&	67.10&	60.85&	75.16&	71.22&	68.66&	64.47&	66.55 \\ 
        BERT\ba-whitening & 57.83& 66.90 & 60.90 & 75.08& 71.31& 68.24& 63.73& 66.28\\ 
        IS-BERT\ba & 56.77 & 69.24 & 61.21 & 75.23 & 70.16 & 69.21 & 64.25 & 66.58 \\
        CT-BERT\ba & 61.63 & 76.80 & 68.47 & 77.50 & 76.48 & 74.31 & 69.19 &72.05 \\
        ConSERT\ba & 64.64 & 78.49 & 69.07 & 79.72 & 75.95 & 73.97 & 67.31 & 72.74 \\
        SimCSE-BERT\ba & 68.40&	82.41 &	74.38 & 80.91 &	78.56 &	76.85 &	72.23 &	76.25 \\
        DCLR-BERT\ba & 70.81 & 83.74 & 75.11 & 82.56 & 78.44 & 78.31 & 71.59 & 77.22 \\
        ArcCSE-BERT\ba & 72.08 & 84.27 & 76.25 & 82.32 & 79.54 & 79.92 & \tf{72.39} & 78.11 \\
        
        $*$ \tf{\ours-BERT\ba} & \tf{74.03} & \tf{84.90} & \tf{76.40} & \tf{83.40} & \tf{80.23} & \tf{81.14} & 71.33 & \tf{78.78} \\
        \midrule
        ConSERT\la & 70.69 & 82.96 &74.13 & 82.78 & 76.66 & 77.53 & 70.37 & 76.45 \\
        SimCSE-BERT \la & 70.88 & 84.16 & 76.43 & 84.50 & 79.76 & 79.26 & 73.88 & 78.41 \\
        DCLR-BERT \la & 71.87 & 84.83 & 77.37 & 84.70 & 79.81 & 79.55 & 74.19 & 78.90 \\
        ArcCSE-BERT\la & 73.17 & \tf{86.19} & \tf{77.90} & \tf{84.97} & 79.43 & 80.45 & 73.50 & 79.37 \\
        $*$ \tf{\ours-BERT\la} & \tf{74.65} & 85.79 & 77.49 & 84.71 & \tf{80.33} & \tf{81.48} & \tf{75.34} & \tf{79.97} \\
        
        
        \midrule
        RoBERTa\ba~(first-last avg.) & 40.88&	58.74&	49.07&	65.63&	61.48&	58.55&	61.63&	56.57\\
        RoBERTa\ba-whitening & 46.99& 63.24&	57.23&	71.36&	68.99&	61.36&	62.91& 61.73\\ 
        DeCLUTR-RoBERTa\ba & 52.41 & 75.19& 65.52 & 77.12 & 78.63 & 72.41 & 68.62 & 69.99\\
        SimCSE-RoBERTa\ba & 70.16 & 81.77 & 73.24 & 81.36 & 80.65 & 80.22 & 68.56 & 76.57 \\
        DCLR-RoBERTa\ba & 70.01 &83.08 & 75.09 & \tf{83.66} & 81.06 & \tf{81.86} & 70.33 & 77.87 \\
        $*$ \tf{\ours-RoBERTa\ba} & \tf{70.73} &\tf{83.77}	& \tf{75.56}	& 81.85	& \tf{83.25}	& 81.43	& \tf{70.96} & \tf{78.22} \\
    \bottomrule
    \end{tabular}
    \end{center}

    \caption{
        Sentence embedding performance on STS tasks. We use the Spearman correlation to measure the relevance between gold annotations and scores predicted by sentence representations, we show the best performance in bold.
    }
    \label{tab:main_sts}
    \vspace{-5pt}
\end{table*}

%% file: tables/main_transfer.tex

\begin{table*}[ht]
    \begin{center}
    \centering
    \small

    \begin{tabular}{lcccccccc}
    \toprule
       \tf{Model} & \tf{MR} & \tf{CR} & \tf{SUBJ} & \tf{MPQA} & \tf{SST} & \tf{TREC} & \tf{MRPC} & \tf{Avg.}\\
    \midrule
    \midrule
        GloVe (avg.) & 77.25&    78.30&  91.17&  87.85&  80.18&  83.00& 72.87 & 81.52\\
        Skip-thought &  76.50& 80.10&  93.60&  87.10&  82.00&  92.20&  73.00& 83.50  \\
        Avg. BERT embeddings & 78.66 & 86.25 & 94.37 & 88.66 & 84.40 & 92.80 & 69.54 & 84.94 \\
        BERT-\cls embedding & 78.68 & 84.85 & 94.21 & 88.23 & 84.13 & 91.40 & 71.13 & 84.66 \\
        IS-BERT\ba & 81.09 & 87.18 & 94.96 & 88.75 & 85.96 & 88.64 & 74.24 & 85.83 \\
        SimCSE-BERT\ba & 81.18 & \tf{86.46} & 94.45 & 88.88 & 85.50 & \tf{89.80} & 74.43 & 85.81 \\
        MoCoSE-BERT\ba & 81.07 & 86.43 & 94.76 & 89.70 & \tf{86.35} & 84.06 & \tf{75.86} & 85.46 \\
        ArcCSE-BERT\ba & 79.91 & 85.25 & \tf{99.58} & 89.21 & 84.90 & 89.20 & 74.78 & 86.12 \\
        $*$ \ours-BERT\ba & \tf{81.31} &86.33 &96.15 &\tf{89.78} &86.08 &89.74 &75.43 &\tf{86.40}\\
        \midrule
        SimCSE-BERT\la & 85.36 & 89.38 & 95.39 & 89.63 & 90.44 & 91.80 & 76.41 & 88.34 \\
        MoCoSE-BERT\la & 83.71 & 89.07 & 95.58 & \tf{90.26} & 87.96 & 84.92 & \tf{76.81} & 86.90 \\
        ArcCSE-BERT\la & 84.34 & 88.82 & \tf{99.58} & 89.79 & 90.50 & 92.00 & 74.78 & 88.54 \\
        $*$ \ours-BERT\la & \tf{85.54} &\tf{89.70} & 96.16 &89.57 &\tf{90.74} &\tf{92.21}	&76.78 & \tf{88.67} \\

    \bottomrule
    \end{tabular}
    \end{center}

    \caption{
        Sentence embeddings performance on transfer tasks. We use the accuracy to measure the performance, and report the best in bold.
    }
    \label{tab:main_transfer}
\end{table*}

%% file: figures/alignment_and_uniformity.tex
\begin{figure}[h]
    \centering 
    \subfigure[alignment]{
        \label{subfig:alignment}
        \includegraphics[width=0.45\columnwidth]{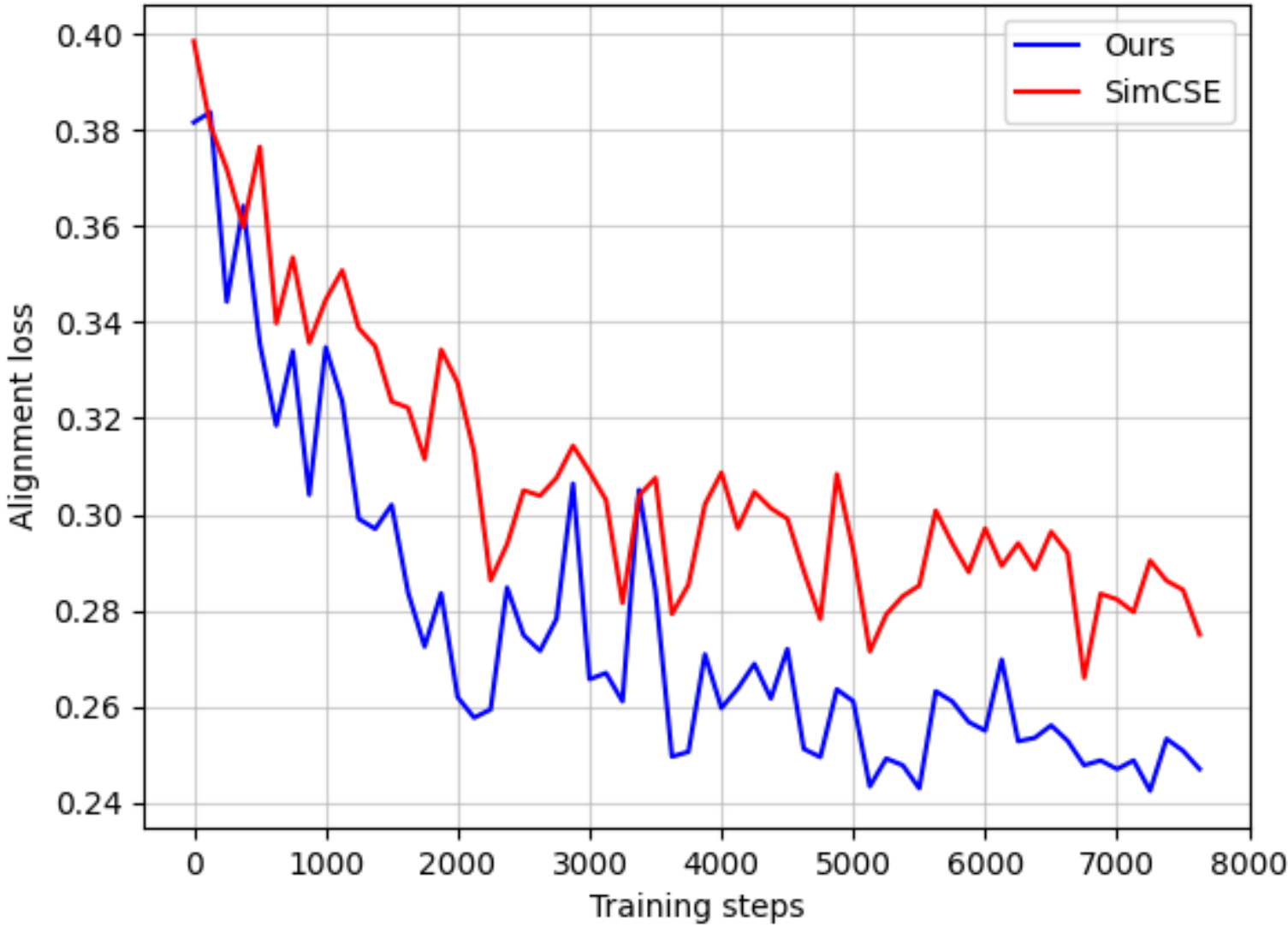}
    }\subfigure[uniformity]{
        \label{subfig:uniformity}
        \includegraphics[width=0.45\columnwidth]{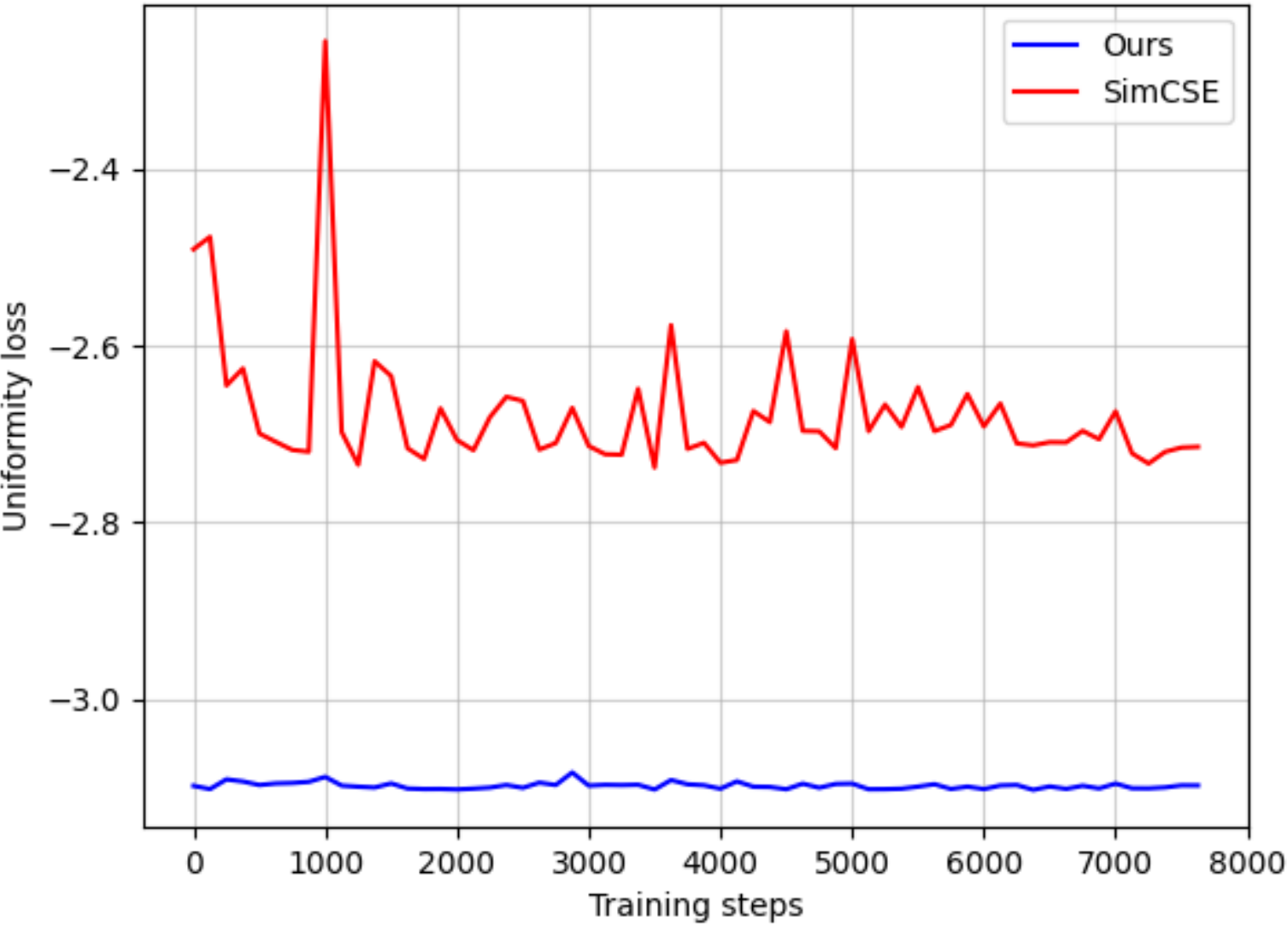}
    }
    \caption{The alignment loss and the uniformity loss of \ours and SimCSE. For both measures, lower number are better.}
    \label{fig:alignment_and_uniformity}
\end{figure}

%% file: sections/analysis.tex
\section{Ablation Analysis}
In this section, we further investigate the effectiveness of our proposed model \ours. For all experiments, we use BERT\ba~ as our base model, and evaluate \ours on the STS tasks unless otherwise specified.

\subsection{Shuffling augments the positive samples}
We prove theoretically and practically that SGW can be regarded as an effective data augmentation. We know different whitening transformations will get different whitened results, but all of them are representations for the same sample, so they can be regarded as positive samples for each other. In \ours, we operate randomly shuffled on feature dimension, and divide the representations along the feature dimension into $k$ groups. Since each time we use a different permutation, we can get a different representations $\mf{Z}$ and the corresponding whitening matrix $\mf{W}$, We find it can be written as a form of feature-wise disturbance $\mf{Z^*} = \mf{Z}+\epsilon$:
\begin{equation}
    \label{eq:augmentation}
    \begin{aligned}
        \mf{Z^*} &= \mf{Z} + \mf{(W^{ZCA} - 1)Z} \\
        &= \mf{Z} + (\mathbf{U \Lambda^{-1/2} U^{\mathrm{T}}}-1)\mf{Z} 
    \end{aligned}
\end{equation}
Here, we treat $(\mathbf{U \Lambda^{-1/2} U^{\mathrm{T}}}-1)\mf{Z}$ as a perturbation $\epsilon$ on the feature dimension. Thus, in \ours, we use it as a data augmentation and generate more diverse positive samples. From Tab.\ref{tab:group_size}, we can see that, shuffling plays a very important role in the performance of the model. 

\subsection{The importance of Group Whitening}
Recently, ~\citet{su2021whitening} directly apply whitening on the output of the BERT and have achieved remarkable performance at the time. 
This lead us to think whether whitening can be directly applied to the output of contrastive learning model to further improve the uniformity of the model representation space.

We consider two different whitening methods: PCA Whitening and ZCA Whitening. The difference between them is that ZCA Whitening uses an additional rotation matrix to rotate the PCA whitened data back to the original feature space, which can make the transformed data closer to the original input data.
\begin{equation}
    \label{eq:PCA_and_ZCA}
    \begin{aligned}
        \mathbf{W}^{ZCA} = \mathbf{U}_{rotate}\mathbf{W}^{PCA}
    \end{aligned}
\end{equation}

\input{tables/whitening.tex}

We use the in-batch sentence representations to calculate the mean value $\bar{x}$ and the covariance matrix $\sigma$, and use the momentum to estimate the overall mean value $\mu$ and covariance matrix $\Sigma$. 
\begin{equation}
    \label{eq:momentum}
    \begin{aligned}
        \mu_n &= \beta \mu_{n-1} + (1-\beta) \bar{x}_{n-1} \\
        \Sigma_n &= \beta \Sigma_{n-1} + (1-\beta) \sigma_{n-1} \\
    \end{aligned}
\end{equation}

As the results shown in the Tab.~\ref{tab:whitening}, we found that directly applying the whitening transformation on contrastive learning models is detrimental to the performance. we attribute this to two reasons: (1) small batch size may not provide enough samples to obtain a suitable estimate for the full covariance matrix. (2) The covariance matrix obtained by high-dimensional features is not necessarily a positive definite matrix (maybe a semi-positive definite matrix), which may leads to errors in matrix decomposition. To alleviate this problem, we use the group whitening to control the extent of whitening. From Tab. \ref{tab:whitening} we can see that group whitening can significantly improve the performance.

\subsection{Hyperparameters Analysis}
For hyperparameters analysis, we want to explore the sensitivity of \ours to these parameters. Concretely, we study the impact of the group size, the number of positive samples. We evaluate our model with varying values, and report the performances on the seven STS tasks. 

\input{tables/group_size.tex}
\paragraph{The influence of group size.}
In \ours, we divide the representation into $k$ groups. However, we know the size of group controls the degree of whitening, and  has a great effect on the effectiveness of \ours, so we carry out an experiment with $k$ varying from 32 to 384. 
As shown in Tab.~\ref{tab:group_size}, we can see that the best performance is achieved when $k=384$, and the second best performance is achieved when $k=128$. When $k$ takes other values, the performance will drop slightly.

\input{tables/positive_num.tex}
\paragraph{The influence of positive samples number.}
Sampling multi-positive samples can significantly enrich semantic diversity, we want to explore how the number of positive samples affect the performance of our model, so we conduct an experiment with positive number $m$ varying from 2 to 5. From Tab.~\ref{tab:positive_num}, we can see that when $m=3$, our model achieve the best performance. However, due to the limitation of the memory size, we cannot exhaust all the possibilities, but we found that when $m \geq 2$, the performance of the model is always better than when $m=2$, which confirms that mult-positive samples can bring richer semantics, allowing the model to learn better sentence representations.

\paragraph{The influence of different modules.}
In \ours, using the proposed shuffled-group-whitening method is beneficial, and further using multi-hot positive samples brings additional benefit. We investigate their respective contributions to \ours in Tab.~\ref{tab:independntly}. In our ablation study, we replace the whitening method with ordinary dropout technique, and still retain the multi-hot positive sample loss. Meanwhile, we use the proposed whitening method alone, and keep the number of positive samples 2. From Tab.~\ref{tab:independntly}, we find that multi-hot positive samples based on dropout only brings +0.12\% improvement. This is reasonable because when the data augmentation is subtle (i.e., the dropout), using extra positive samples barely increases the diversity. In contrast, the proposed SGW generates informative data augmentation, and thus well accommodates the multi-hot positive samples.

\input{tables/ablation}

%% file: tables/whitening.tex
\begin{table}[t]
    \begin{center}
    \centering
    \small
    \resizebox{0.88\columnwidth}{!}{%
    \begin{tabular}{lccc}
    \toprule
    \tf{Whitening strategy} & & & \tf{STS-Avg.}\\
    \midrule
    None (unsup. SimCSE) & & & 76.25 \\
    \midrule
        PCA-whitening & & &  68.55 \\
        ZCA-whitening &  &  &  72.11 \\
        Group-whitening & & & 77.47 \\
        Shuffled-Group-whitening & & & \tf{78.78} \\
    \bottomrule
    \end{tabular} 
    }
    \end{center}

    \caption{
        Performance comparison using different whitening methods on the test set of seven semantic textual similarity tasks. (Spearman's correlation).
    }
    \label{tab:whitening}
\end{table}

%% file: tables/group_size.tex
\begin{table}[t]
    \begin{center}
    \centering
    \small
    \resizebox{0.8\columnwidth}{!}{%
    \begin{tabular}{lcccc}
    \toprule
       \tf{Group Size} & \emph{32} & \emph{48} & \emph{64} & \emph{96}\vspace{2pt}  \\
       \ti{STS-Avg.} &76.49 &77.59 & 78.34 & 78.03 \\
       \ti{w/o shuffled} &75,49 & 77.47  &77.42  & 77.45  \\
       \midrule
       \tf{Group Size} & \emph{128} &\emph{192} & \emph{256} & \emph{384}\vspace{2pt} \\
       \ti{STS-Avg.} & \underline{78.57} & 78.15 & 77.97 & \tf{78.78}\\
       \ti{w/o shuffled} &77.47  &77.45  & 77.42  & 77.43 \\
    \bottomrule
    \end{tabular}
    }
    \end{center}

    \caption{
        Effects of different group size and shuffling on seven STS tasks (Spearman's correlation).The best performance and the second-best performance methods are denoted in bold and underlined fonts respectively.
    }
    \label{tab:group_size}
    \vspace{-4pt}
\end{table}

%% file: tables/positive_num.tex
\begin{table}[t]
    \begin{center}
    \centering
    \small
    \resizebox{0.8\columnwidth}{!}{%
    \begin{tabular}{lcccc}
    \toprule
       \tf{Positive Number} & \emph{2} & \emph{3} & \emph{4} & \emph{5}\vspace{2pt}  \\
       \midrule
       \ti{STS-Avg.} & 77.81 & \tf{78.78} &  78.65  & 78.74\\
    \bottomrule
    \end{tabular}
    }
    \end{center}

    \caption{
        Effects of different positive number on seven STS tasks (Spearman's correlation).
    }
    \label{tab:positive_num}
    \vspace{-4pt}
\end{table}

%% file: tables/ablation.tex
\begin{table}[t]
    \begin{center}
    \centering
    \small
    \resizebox{0.88\columnwidth}{!}{%
    \begin{tabular}{lccc}
    \toprule
    \tf{Model} & & & \tf{STS-Avg.}\\
    \midrule
    SimCSE-BERT-base & & & 76.25 \\
    \midrule
    ~ + Shuffled Group Whitening & & & 77.81~(+1.56)\\
    \midrule 
    ~ + Multiple Positives(k=3) & & & 76.37~(+0.12) \\
    ~ + Multiple Positives(k=4) & & & 76.34~(+0.09) \\
    \midrule
    \ours-BERT\ba & & & \tf{78.78}~(+2.53) \\
    \bottomrule
    \end{tabular}
    }
    \end{center}

    \caption{
        The performance on STS tasks when we use whitening module and multiple positives independently.
    }
    \label{tab:independntly}
\end{table}

%% file: sections/conclusion.tex
\section{Conclusion}

In this paper, we proposed \ours, a whitening-based contrastive learning framework for unsupervised sentence representation learning. We proposed a novel shuffled group whitening, which reinforces the contrastive learning effect regarding both the uniformity and the alignment. Specifically, it retains the role of whitening in dispersing data, and can further improve uniformity on the basis of contrastive learning. Additionally, it shuffles and groups features on channel axis, and performs whitening independently within each group. This kind of operation can be regarded as a disturbance on feature dimension. We obtain multiple positive samples through this operation, and learn the invariance to this disturbance to obtain better alignment. Experimental results on seven semantic textual similarity tasks have shown that our approach achieve consistent improvement over the contrastive learning baseline.                                                                                                                                                                                                                                 

%% file: sections/limitation.tex
\section*{Limitations}

In this paper, we limit the proposed \ours for sentence embedding learning. Conceptually, \ours is potential to benefit contrastive learning on some other tasks, \emph{e.g.}, self-supervised image representation learning and self-supervised vision-language contrastive learning. However, we did not investigate the self-supervised image representation learning because this domain is currently dominated by masked image modeling. We will consider extending \ours for vision-language contrastive learning when we have sufficient training resources for the extraordinary large-scale text-image pairs. 

\section*{Ethics Statement}

This paper is dedicated to deep sentence embedding learning and proposes a new unsupervised sentence representation model, it does not involve any ethical problems. In addition, we are willing to open source our code and data to promote the better development of this research direction

\section*{Acknowledgements}

This work is supported by National Natural Science Foundation of China (No. 62293554) and the Fundamental Research Funds for the Central Universities (No. 226-2022-00051).